\documentclass[letterpaper]{article} 
\usepackage{aaai2026}  
\usepackage{times}  
\usepackage{helvet}  
\usepackage{courier}  
\usepackage[hyphens]{url}  
\usepackage{graphicx} 
\usepackage{natbib}  
\usepackage{caption} 
\usepackage{algorithm}
\usepackage{algorithmic}

\usepackage{newfloat}
\usepackage{listings}

\usepackage{booktabs, multicol, multirow, amsmath, amssymb}

\usepackage{etoolbox}

\AtBeginEnvironment{equation}{\small}
\DeclareCaptionStyle{ruled}{labelfont=normalfont,labelsep=colon,strut=off} 
\floatstyle{ruled}
\newfloat{listing}{tb}{lst}{}
\floatname{listing}{Listing}

\title{Rethinking Target Label Conditioning in Adversarial Attacks: A
	2D Tensor-Guided Generative Approach}
\author{
    Hangyu Liu\textsuperscript{\rm 1,\rm 2},
    Bo Peng\textsuperscript{\rm 3},
    Pengxiang Ding\textsuperscript{\rm 1,\rm 2},
    Donglin Wang\textsuperscript{\rm 2}\thanks{Corresponding author.}
}
\affiliations{
    \textsuperscript{\rm 1}Zhejiang University\\
    \textsuperscript{\rm 2}Westlake University\\
    \textsuperscript{\rm 3}Beijing University of Posts and Telecommunications\\

    12563051@zju.edu.cn, bonovice@bupt.edu.cn,  dingpengxiang@westlake.edu.cn, wangdonglin@westlake.edu.cn
}

\begin{document}

\maketitle

\begin{abstract}
Compared to single-target adversarial attacks, multi-target attacks have garnered significant attention due to their ability to generate adversarial images for multiple target classes simultaneously. However, existing generative approaches for multi-target attacks primarily encode target labels into one-dimensional tensors, leading to a loss of fine-grained visual information and overfitting to model-specific features during noise generation. To address this gap, we first identify and validate that the semantic feature quality and quantity are critical factors affecting the transferability of targeted attacks: 1) Feature quality refers to the structural and detailed completeness of the implanted target features, as deficiencies may result in the loss of key discriminative information; 2) Feature quantity refers to the spatial sufficiency of the implanted target features, as inadequacy limits the victim model's attention to this feature. Based on these findings, we propose the 2D Tensor-Guided Adversarial Fusion (TGAF) framework, which leverages the powerful generative capabilities of diffusion models to encode target labels into two-dimensional semantic tensors for guiding adversarial noise generation. Additionally, we design a novel masking strategy tailored for the training process, ensuring that parts of the generated noise retain complete semantic information about the target class. Extensive experiments demonstrate that TGAF consistently surpasses state-of-the-art methods across various settings.
\end{abstract}

\begin{links}
    \link{Code}{https://github.com/TemenosMistral/TGAF}
\end{links}

\section{Introduction}
\begin{figure}[!t]
	\centering
	\includegraphics[width=\linewidth]{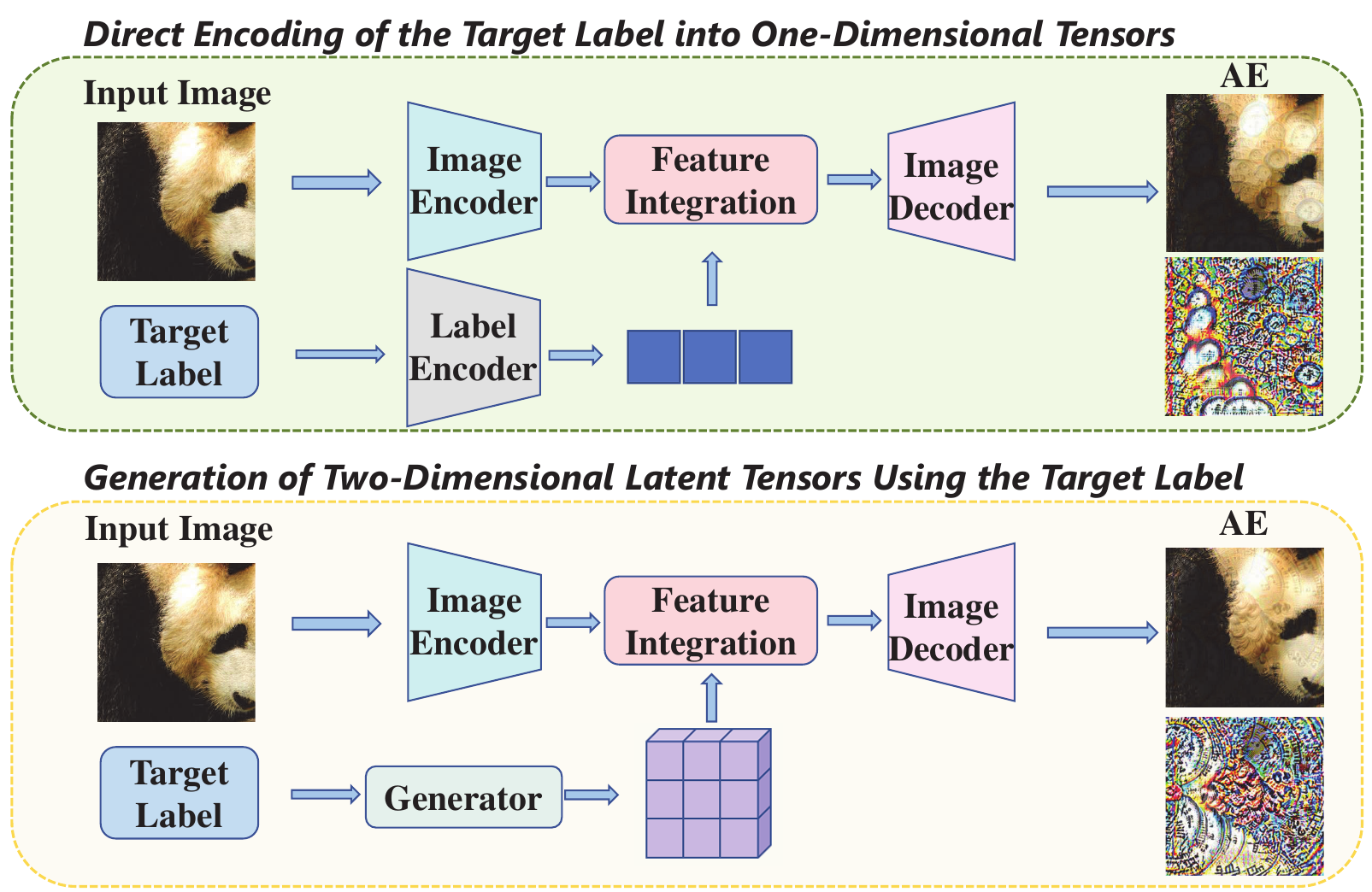}
	\caption{\textbf{Comparison of multi-target approaches.} Previous methods (top row) use 1D label encoding to guide noise generation for adversarial examples (AE). However, this often loses fine-grained details because images are 2D, potentially leading to overfitting. Our method (bottom row) generates 2D latent representations from target labels, better preserving structural information.}
	\label{fig:teaser}
\end{figure}

\begin{figure*}[!t]
	\centering
	\includegraphics[width=0.65\textwidth]{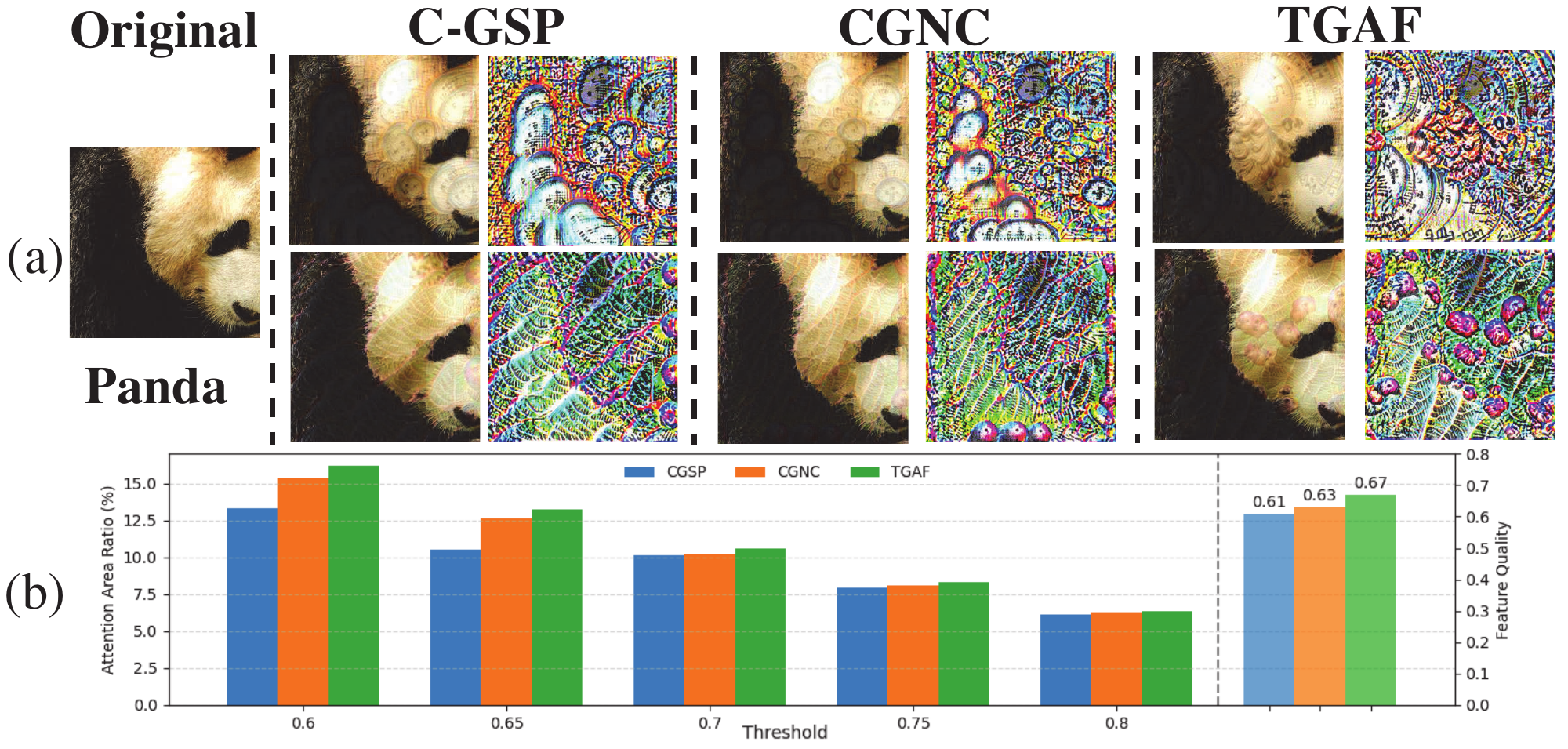}
	
    \caption{
    \textbf{(a) Visualization comparison of C-GSP, CGNC, and TGAF.} Each row displays an adversarial example and its corresponding perturbation map for a distinct target label: ``barometer'' (first row) and ``fig'' (second row). The surrogate model used is Inc-v3. TGAF demonstrably surpasses C-GSP and CGNC by more effectively capturing both the target semantic details (e.g., the barometer's pointer) and the target semantic quantity (e.g., the number of figs). \textbf{(b) Quantitative analysis of feature quantity and feature quality.} Feature quantity is measured by the percentage of high-attention area analyzed via Grad-CAM. Feature quality is measured by the cosine similarity between the perturbation's feature vector and the average feature vector of real target class images. Experiments are conducted on 1000 images.
}
	\label{fig:motivation}
\end{figure*}

With the rapid advancement of deep neural networks (DNNs), artificial intelligence has achieved significant progress in fields such as medical diagnostics~\cite{ma2021understanding}, image classification~\cite{he2016deep}, and autonomous driving~\cite{kong2020physgan}. However, the adversarial vulnerability of deep learning models has emerged as a major challenge affecting their reliable application in security-sensitive scenarios.

Adversarial attacks manipulate input data by introducing subtle and imperceptible perturbations, misleading model predictions. These attacks can be categorized into two main types: untargeted attacks, which aim to cause misclassification without specifying a particular category, and targeted attacks, which force models to output labels chosen by the attacker. Targeted attacks are not only more challenging but also more threatening as they enable specific malicious control, such as deceiving an autonomous vehicle into interpreting a stop sign as a speed limit sign.

In real-world applications, attackers typically do not fully understand the architecture or parameters of the target model, making transferability a crucial property for effective adversarial attacks. Transferability refers to the ability of adversarial examples crafted on surrogate models to successfully deceive unknown black-box models~\cite{wang2021enhancing}. While untargeted attacks have demonstrated strong transferability, targeted attacks still suffer from low success rates due to their dependency on overfitting the decision boundaries of surrogate models.

Existing targeted attack methods primarily fall into two categories: instance-specific approaches~\cite{dong2018boosting,gao2021feature} and instance-agnostic approaches~\cite{feng2023dynamic,kong2020physgan}. Instance-specific methods~\cite{dong2018boosting,dong2019evading} optimize adversarial perturbations for individual samples in an iterative manner, often leading to inefficiency and susceptibility to overfitting. On the other hand, instance-agnostic approaches focus on learning universal perturbations~\cite{moosavi2017universal,li2022learning} or generators~\cite{naseer2019cross} based on data
distribution to enhance attack generality. Recent generative techniques, such as C-GSP and CGNC, have significantly improved efficiency by training conditional generators for multiple target classes compared to single-target methods. However, these methods simply use 1D tensors to guide generation and fail to thoroughly investigate the key factors affecting attack transferability.

\begin{figure}[!t]
	\centering
	\includegraphics[width=\linewidth]{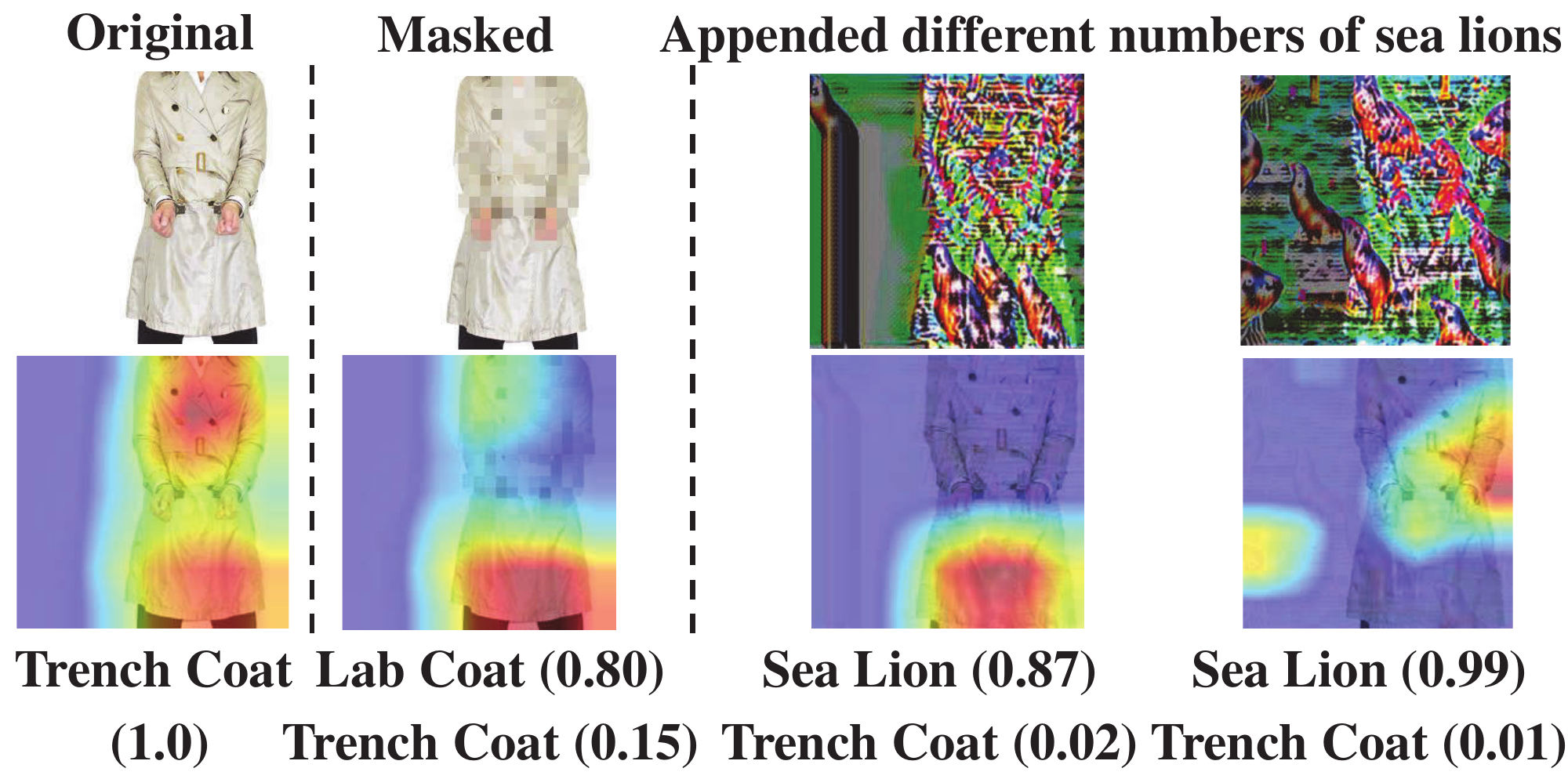}
	\caption{\textbf{Demonstrating the impact of feature implantation.} The heatmaps are generated on Res-152 using Grad-CAM. Left: The original image and heatmap. Middle: The masked image and heatmap (Masked trench coat buttons). Right: The perturbation and heatmap.  Results confirm that 1) insufficient feature details reduce accuracy, while 2) more features have a higher target probability.}
	\label{fig:suppose1}
\end{figure}

To address this gap, we observe that adversarial noise generated for targeted attacks functions similarly to ``implanting'' semantic features of the target class into source images. Building on this insight, we define two critical factors that influence attack transferability: the quality and quantity of target-class semantic features in the generated noise. Through experimental analysis (Fig.~\ref{fig:suppose1}), we validate our hypothesis. Furthermore, we examine the adversarial noise produced by C-GSP and CGNC and identify issues related to incomplete semantic feature implantation and insufficient semantic representation, as shown in Fig.~\ref{fig:motivation}. We hypothesize that these shortcomings arise because these methods represent target labels as 1D tensors (e.g., one-hot encoding or CLIP embeddings), disregarding spatial and structural information, and tend to overfit to the regions where noise mapping is easiest during training. Consequently, the generated adversarial noise lacks essential target-class features and sufficient semantic information, thereby limiting its transferability.

To overcome these limitations, we propose a novel generative framework termed 2D-Tensor-Guided Adversarial Fusion (TGAF). The core innovation of our approach lies in leveraging 2D spatial information of target labels to guide adversarial noise generation. This enables the retention of fixed low-level semantic information during the generation process, rather than relying solely on the decision boundaries of surrogate models. Additionally, we introduce a carefully designed random masking strategy tailored for training, ensuring that parts of the generated noise still contain complete semantic information of the target class. Our contributions can be summarized as follows:

\begin{itemize}
	\item We first systematically analyze adversarial noise generation from the perspective of semantic feature quality and quantity, uncovering their impact on transferability.
	
	\item  We propose TGAF, an innovative approach that combines 2D target representation and random masking strategies to enhance adversarial transferability.
	
	\item Our method significantly outperforms previous attack methods in terms of targeted transferability across various experimental settings.
\end{itemize}

\section{Related Work}

\subsection{Transferable Targeted Attacks}

Transferable targeted adversarial attacks aim to fool multiple models into misclassifying adversarial examples into a specific target class. To enhance attack transferability, input transformation-based methods ~\cite{byun2022improving,wei2023enhancing} diversify input representations through techniques such as geometric transformations and local mixup. Advanced objective-based approaches~\cite{weng2023logit,byun2023introducing} refine loss functions or optimize specific outputs. Generation-based methods~\cite{naseer2021generating,zhao2023minimizing} employ generative models to produce adversarial examples that closely resemble the target class characteristics and ensemble-based techniques~\cite{wu2024improving} focus on efficiently integrating multiple models or strategies, leveraging approaches like self-distillation and weight scaling.

\subsection{Defense Methods}

To mitigate adversarial attacks, various defense strategies have been developed. Among them, Adversarial Training (AT)~\cite{goodfellow2014explaining} is one of the most effective, improving robustness by incorporating adversarial examples during training. Preprocessing-based defenses~\cite{dziugaite2016study,xu2017feature} instead remove adversarial perturbations before inference, while denoising techniques like HGD~\cite{liao2018defense} use autoencoders guided by high-level features to purify inputs. Diffusion purification~\cite{naseer2020self,nie2022diffusion} further enhances defense by leveraging diffusion models to reconstruct clean images.

\section{Methodology}

\begin{figure*}[!t]
	\centering
	\includegraphics[width=0.7\textwidth]{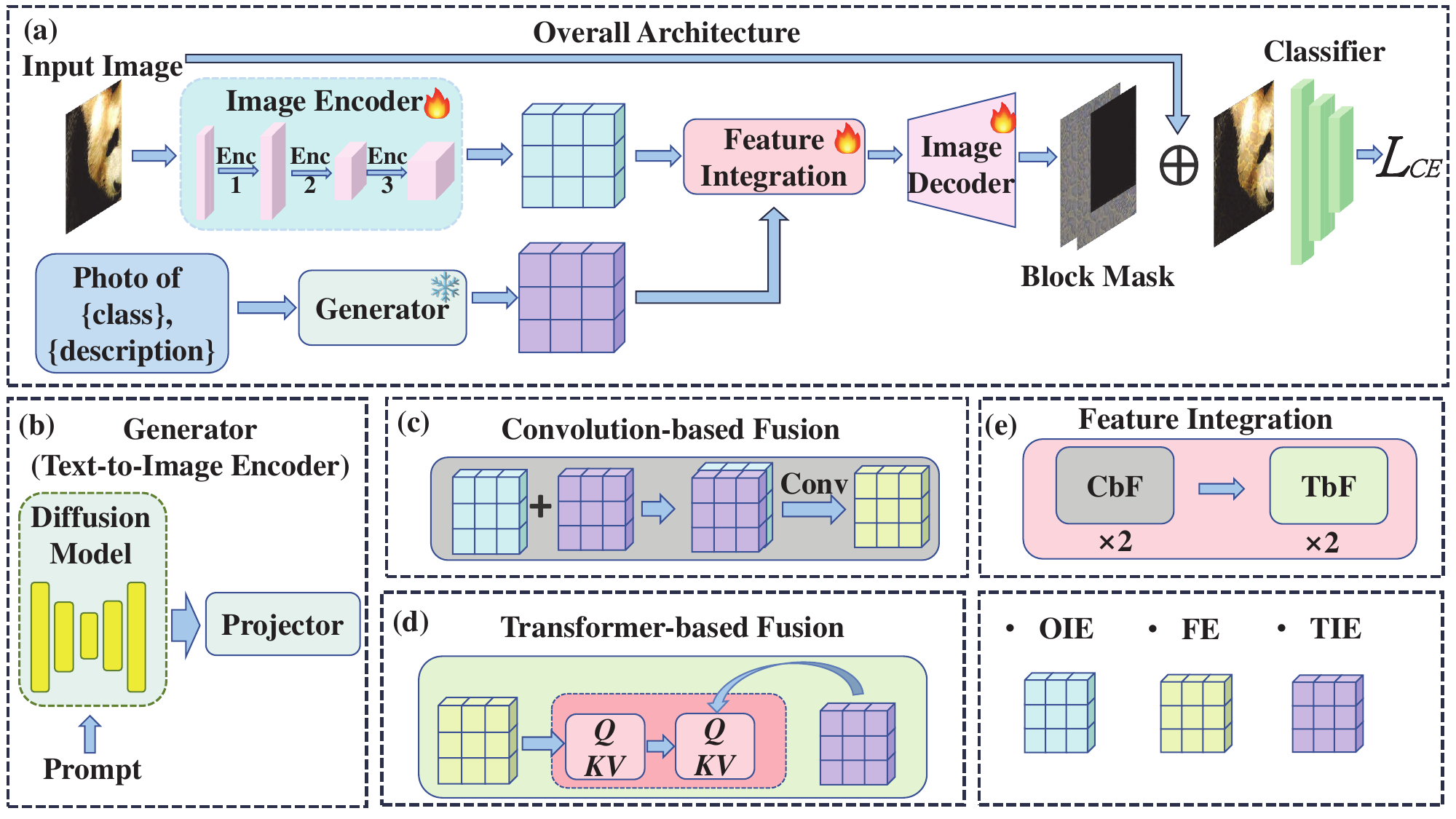}
	\caption{
		\textbf{The framework of 2D Target-Guided Adversarial Fusion (TGAF).} (a) Overall architecture; (b) Text-to-Image Encoder; (c) Convolution-based Fusion (CbF) module; (d) Transformer-based Fusion (TbF) module; (e) Feature Integration
		Module. Abbreviations: OIE (Original Image Embeddings), FE (Fusion Embeddings), TIE (Target Image Embeddings).
    }
	\label{fig:framework}
\end{figure*}

In this section, we first introduce the prerequisite knowledge and our motivation. Then, we provide a detailed explanation of our TGAF method.

\subsection{Preliminaries}
Let \( f_\Phi: \mathcal{X} \rightarrow \mathcal{Y} \) denote a white-box image classifier parameterized by \(\Phi\), where \(\mathcal{X} \in \mathbb{R}^{C \times H \times W}\) represents the image domain and \(\mathcal{Y} \in \mathbb{R}^L\) denotes the output confidence scores over \(L\) classes. Given a natural image \( x \in \mathcal{X} \) and a target class \( c_t \in \mathcal{C} \), the goal of transferable targeted adversarial attacks is to craft an imperceptible perturbation \(\delta\) such that the adversarial example \( x_{adv} = x + \delta \) misleads both the surrogate model \( f_\Phi \) and unknown black-box victim models \( F_\Phi \) into predicting \( c_t \). This objective can be formalized as follows:
\begin{equation}
	\arg\max f_\Phi(x_{adv}) = c_t, \quad \text{with} \quad \|\delta\|_\infty \leq \epsilon,
\end{equation}
where \(\epsilon\) constrains the perturbation magnitude under the \( l_\infty \)-norm.

To achieve this objective, numerous instance-specific methods have been proposed. However, these approaches often require a large number of iterations per attack, resulting in high computational cost and low efficiency. To address this limitation, several multi-target methods have been developed, primarily based on a generative framework.

Existing generative approaches train a conditional perturbation generator \( G_{\theta} \) to map input pairs \((x, c_t)\) to targeted perturbations, i.e., \(\delta = G_{\theta}(x, c_t)\). The optimization objective is formulated as:
\begin{equation}
	\begin{gathered}
		\min_{\theta} \; \mathbb{E}_{x \sim \mathcal{X},\, c_t \sim \mathcal{C}} \Bigl[\,\mathcal{L}\bigl(f_\Phi\bigl(x + G_{\theta}(x, c_t)\bigr), c_t\bigr)\Bigr], \\
		\text{s.t.} \quad \bigl\|G_{\theta}(x, c_t)\bigr\|_{\infty} \le \varepsilon.
	\end{gathered}
\end{equation}
Here, \(\mathcal{X}\) represents an unlabeled training set and \(\mathcal{L}(\cdot, \cdot)\) is typically the cross-entropy loss.


\subsection{Motivation}

Although recent advances in targeted adversarial attacks have been made, the transferability of generated adversarial samples remains unsatisfactory. In non-targeted adversarial attacks, numerous studies~\cite{wang2024boosting,liu2025enhancing} have highlighted that different models focusing on distinct object regions impede the generalization of adversarial samples. This observation naturally leads us to investigate:

\emph{What primarily hinders the generalizability of adversarial samples in targeted attacks?}

Existing studies~\cite{yang2022boosting,fang2024clip} suggest that targeted attacks achieve their objective by ``implanting'' target-class semantic features into source images. Based on this, we hypothesize that the nature of these implanted features is the key factor influencing attack success. Specifically, we examine: 1) \textbf{Feature quality}: The integrity and detail preservation of implanted features. 2) \textbf{Feature quantity}: The number of target-class features introduced into the source image. As shown in Fig.~\ref{fig:suppose1}, our experiments reveal two critical phenomena: 1) When essential details (e.g., buttons on a trench coat) are masked, models exhibit high-confidence misclassifications (e.g., predicting ``lab coat'' instead of the true class). 2) Increasing the number of implanted target features (e.g., sea lion patches)  raises the likelihood of target-class predictions. These two factors jointly affect adversarial sample generalizability. Since different models focus on disparate object regions, feature omissions may prevent certain models from capturing their critical discriminative features, while insufficient feature quantity further constrains attack effectiveness.

Based on our previous findings, we analyzed the adversarial noise generated by different multi-target approaches. As shown in Fig.~\ref{fig:motivation} (a), we found that: 1) for the barometer class, the adversarial noise produced by C-GSP and CGNC methods lacks essential structural features including the pointer and scale markings; 2) for the fig class, these methods generate noise that primarily concentrates at image boundaries. These findings prompt a novel question: 


\emph{How can we create adversarial noise that incorporates a greater quantity and higher completeness of target features?}

To address this issue, we first conducted an analysis of existing approaches. Fig.~\ref{fig:teaser} shows that current methods universally encode target labels as 1D tensors for processing. However, this 1D encoding scheme fundamentally differs from the 2D nature of images, resulting in the loss of low-level visual information. Such information deficiency introduces bias during training: the models tend to learn specific feature representations from surrogate models, consequently leading to overfitting.

Regarding the issue of insufficient feature injection,  we developed an innovative masked training mechanism tailored for training inspired by CGNC's fine-tuning approach. This mechanism randomly masks the model's output noise during training, ensuring that partial noise regions retain complete feature information.

\subsection{2D Target-Guided Adversarial Fusion}

The architecture of TGAF method is illustrated in Fig.~\ref{fig:framework}. Specifically, TGAF comprises four key components: an Image Encoder $\mathcal{E}$, a Text-to-Image Encoder $\mathcal{G}$, a Feature Integration Module $\mathcal{F}$, and an Image Decoder $\mathcal{D}$. We next detail the design of each component.

\textbf{Image Encoder module.} Given an input image $\mathbf{I} \in \mathbb{R}^{C \times H \times W}$, our encoder $\mathcal{E}$ extracts multi-scale features through three convolutional layers, gradually abstracting and downsampling the features. Specifically, we transform the image features from $C \times H \times W$ to $C' \times \frac{H}{4} \times \frac{W}{4}$. Let $\mathbf{x} \in \mathbb{R}^{C' \times \frac{H}{4} \times \frac{W}{4}}$ denote the encoded image representation.

\textbf{Text-to-Image Encoder Module.} To obtain a 2D representation of the target class, we leverage the  capabilities of diffusion models, employing the encoder and denoising Unet parts of Stable-Diffusion-2~\cite{Rombach_2022_CVPR} to process the input target labels, resulting in low-dimensional latent vectors of size $ B \times 4 \times 64 \times 64 $. To align with the features of the original image, we use  a convolutional layer and an  average pooling to project the latent vectors from $ 4 \times 64 \times 64 $ to a latent representation $\mathbf{z}_{c} \in \mathbb{R}^{4 \times \frac{H}{4} \times \frac{W}{4}}$.

\textbf{Feature Integration Module.} The goal of this module is to effectively fuse the original image representation $\mathbf{x}$ with the target-conditioned representation $\mathbf{z}_{c}$. We propose two fusion strategies:

\textbf{1. Convolution-based Fusion (CbF):}  We employ a learnable $1 \times 1$ convolution  to adaptively reduce channel dimensions and learn local feature interactions.
\begin{equation}
	\mathbf{f}_{c} = \text{Conv}_{1\times1}(\mathbf{x} \parallel \mathbf{z}_{c}),
\end{equation}
where $\text{Conv}_{1\times1}$ projects the concatenated features back to the original channel dimension $C'$ of $\mathbf{x}$. Let $x_c$ denote the output of convolution-based fusion.

\textbf{2. Transformer-based Fusion (TbF):} We leverage attention mechanisms to capture complex global spatial-channel dependencies and model long-range interactions between features. The TbF module comprises three sequential computational stages:

First, we transform the condition features  $z_c$ through a $1 \times 1$ convolutional layer to align the channel of $x_c$:

\begin{equation}
	\mathbf{z}_{t} = \text{Conv}_{1\times1}(\mathbf{z}_{c}).
\end{equation}

Subsequently, we apply channel attention~\cite{hu2018squeeze} to recalibrate feature importance:

\begin{align}
	\text{CHA}(\mathbf{x}) &= \sigma\left(W_2 \cdot \text{ReLU}(W_1 \cdot \text{AvgPool}(\mathbf{x}))\right), \\
	\mathbf{x}_{ca} &= \mathbf{x}_{c} \odot \text{CHA}(\mathbf{x}_{c}).
\end{align}

Finally, we apply a transformer fusion mechanism to the channel-attended features, incorporating condition features:

\begin{table*}[!t]
  \centering
  \small
  \setlength{\tabcolsep}{1mm}
  \begin{tabular}{c|c|c|c|c|c|c|c|c|c|c|c}
    \toprule
    \textbf{Source} & \textbf{Method} & \textbf{Inc-v3} & \textbf{Inc-v4} & \textbf{Inc-Res-v2} & \textbf{Res-152} & \textbf{DN-121} & \textbf{GoogleNet} & \textbf{VGG-16} & \textbf{ViT-B} & \textbf{ViF-S} & \textbf{Swin-T} \\
    \hline
    \multirow{6}{*}{\textbf{Inc-v3}}
     & Logit       & \textbf{99.60*} & 5.80  & 6.50   & 1.70   & 3.00   & 0.80   & 1.50   & 0.59    & 3.09  & 1.96  \\
     & SU          & 99.59* & 5.80  & 7.00   & 3.35   & 3.50   & 2.00   & 3.94   & 0.40    & 1.01  & 0.36  \\
     & Everywhere  & 99.31* & 50.72 & 45.46  & 34.40  & 42.34  & 40.28  & 41.18  & 5.73   & 27.19 & 17.73 \\
    \cline{2-12}
     & C-GSP       & 93.40* & 66.90 & \textbf{66.60} & 41.60  & 46.40  & 40.00  & 45.00  & 21.18 & 40.86 & 21.48 \\
     & CGNC        & 96.03* & 59.41 & 47.98  & 42.50  & 62.91  & 51.36  & 52.63  & 24.81  & 52.43 & 28.16 \\
     & TGAF        & 98.15* & \textbf{72.49} & 63.20  & \textbf{61.94} & \textbf{78.30} & \textbf{66.86} & \textbf{70.64} & \textbf{33.03} &  \textbf{63.27} & \textbf{42.61} \\
    \hline
    \multirow{6}{*}{\textbf{Res-152}}
     & Logit       & 10.10  & 10.70 & 12.80  & 95.70* & 12.70  & 3.70   & 9.20   & 2.09    & 21.25 & 5.19  \\
     & SU          & 12.36  & 11.31 & 16.16  & 95.08* & 16.13  & 6.55   & 14.28  & 2.29    & 19.94 & 4.70  \\
     & Everywhere  & 60.38  & 55.20 & 43.00  & 97.66* & 74.16  & 34.80  & 60.90  & 11.77 &  57.08 & 38.83 \\
    \cline{2-12}
     & C-GSP       & 37.70  & 47.60 & \textbf{45.10} & 93.20* & 64.20  & 41.70  & 45.90  & 22.01   & 36.66 & 21.65 \\
     & CGNC        & 53.39  & 51.59 & 34.18  & 95.85* & 85.60  & 62.27  & 63.36  & 34.81  & 58.68 & 40.84 \\
     & TGAF        & \textbf{62.60} & \textbf{62.44} & 44.02  & \textbf{97.79*} & \textbf{87.90} & \textbf{63.54} & \textbf{65.20} & \textbf{39.64} &  \textbf{63.36} & \textbf{42.84} \\
    \bottomrule
  \end{tabular}
  
  \caption{Attack success rates (\%) for multi-target attacks against \textbf{normally trained} models. * represents white-box attacks.}
  \label{normal}
\end{table*}

\begin{equation}
	\mathbf{f}_{t} = \text{Transformer}(\mathbf{x}_{ca}, \mathbf{z}_{t}).
\end{equation}

Specifically, $\mathbf{f}_{t}$ consists of a self-attention and a cross-attention, the
complete $\mathbf{f}_{t}$ can be formulated as:

\begin{equation}
	\mathbf{f}_{t} = \text{CA}(\text{SA}(\mathbf{x}_{ca}), \mathbf{z}_{t}).
\end{equation}

In the equations above, \textbf{SA} (self-attention) and \textbf{CA} (cross-attention) are defined as:

\textbf{Self-Attention (SA):}
\begin{equation}
    \text{SA}(\mathbf{x}_{ca}) = \text{Softmax}\left(\frac{\mathbf{Q}_{ca}\mathbf{K}_{ca}^T}{\sqrt{d_k}}\right)\mathbf{V}_{ca}.
\end{equation}

\textbf{Cross-Attention (CA):} 
\begin{equation}
    \text{CA}(\mathbf{x}_{sa}, \mathbf{z}_{t}) = \text{Softmax}\left(\frac{\mathbf{Q}_{sa}\mathbf{K}_{z_t}^T}{\sqrt{d_k}}\right)\mathbf{V}_{z_t}.
\end{equation}

Here, $\mathbf{Q}_{ca}, \mathbf{K}_{ca}, \mathbf{V}_{ca}$ are query, key, and value matrices from the input $\mathbf{x}_{ca}$, while $\mathbf{K}_{z_t}$ and $\mathbf{V}_{z_t}$ come from the target-conditioned representation $\mathbf{z}_{t}$.

\textbf{Image Decoder Module.} The decoder $\mathcal{D}$ upsamples features from the Feature Integration Module through transposed convolutions to reconstruct $\boldsymbol{o}$. Following C-GSP~\cite{yang2022boosting}, we apply $\tanh(\cdot)$ projection with budget $\epsilon$ to obtain the final output $\boldsymbol{\delta} = \epsilon \cdot \tanh(\boldsymbol{o})$.


\textbf{Mask Mechanism.} We introduce a dynamic block-wise masking strategy that randomly masks regions in the noise pattern $\delta$. Specifically, we partition the image into $N \times N$ blocks of varying sizes and randomly select two blocks for masking.




\textbf{Training Objectives.} The proposed TGAF framework is trained to minimize the cross-entropy loss between the victim model's prediction and the target class label $c_t$ through an end-to-end optimization process:


\begin{equation}
	\theta^* \leftarrow \arg\min_\theta \mathcal{L}_{\text{CE}}\left(f_\Phi(\boldsymbol{x}_s+\mathcal{D}_\theta(\mathcal{F}_\theta(\mathcal{E}_\theta, \mathcal{G}_\theta), c_t\right).
\end{equation}

where the adversarial sample $\boldsymbol{x}_s + \mathcal{D}_\theta(\cdot)$ is fed into the victim model $f_\Phi$ to compute the cross-entropy loss $\mathcal{L}_{\text{CE}}$ that drives the attack towards the target class $c_t$.

\section{Experiments}

\subsection{Experimental Settings}


		


\begin{table*}[!t]
  \centering
  \small
  \setlength{\tabcolsep}{1mm}
  \begin{tabular}{c|c|c|c|c|c|c|c}
    \toprule
    \textbf{Source} & \textbf{Method} & \textbf{Inc-v3\textsubscript{ADV}} & \textbf{IR-v2\textsubscript{ENS}} &
    \textbf{Res50\textsubscript{SIN}} & \textbf{Res50\textsubscript{IN}} & 
    \textbf{Res50\textsubscript{FINE}} & \textbf{Res50\textsubscript{AUG}}  \\
    \hline
    \multirow{6}{*}{\textbf{Inc-v3}}
     & Logit       & 0.30 & 0.30 & 0.70 & 1.23 & 3.14 & 0.86  \\
     & SU          & 0.49 & 0.41 & 0.84 & 1.75 & 3.55 & 1.04  \\
     & Everywhere  & 0.68 & 1.19 & 4.73 & 27.04 & 39.37 & 18.15 \\
    \cline{2-8}
     & C-GSP       & 20.41 & 18.04 & 6.96 & 33.76 & 44.56 & 21.95  \\
     & CGNC        & 24.30 & 22.51 & 8.88 & 40.81 & 52.13 & 22.83 \\
     & TGAF        & \textbf{39.69} & \textbf{34.86} & \textbf{17.76} & \textbf{64.79} & \textbf{72.36} & \textbf{43.53} \\
    \hline
    \multirow{6}{*}{\textbf{Res-152}}
     & Logit       & 1.15 & 1.18 & 1.65 & 6.70 & 15.46 & 5.93  \\
     & SU          & 2.12 & 1.20 & 1.95 & 7.53 & 21.14 & 6.95  \\
     & Everywhere  & 0.55 & 1.23 & 9.71 & 59.94 & 81.45 & 50.09 \\
    \cline{2-8}
     & C-GSP       & 14.60 & 16.01 & 16.84 & 60.30 & 65.51 & 42.88  \\
     & CGNC        & 22.15 & 26.70 & 29.81 & 79.82 & 84.05 & 63.66 \\
     & TGAF        & \textbf{27.73} & \textbf{32.71} & \textbf{38.07} & \textbf{84.53} & \textbf{88.48} & \textbf{68.63}  \\
    \bottomrule
  \end{tabular}

  \caption{Attack success rates (\%) for multi-target attacks against \textbf{robustly trained} models.}
  \label{robust}
\end{table*}

\begin{table*}[!h]
	\centering

    \small
  \setlength{\tabcolsep}{1mm}
	\begin{tabular}{cccccclcccclc}
		\toprule
		\textbf{Source} & \textbf{Method} 
		& \multicolumn{3}{c}{\textbf{Smoothing}} 
		& \multicolumn{3}{c}{\textbf{JPEG compression}} 
		& \multicolumn{3}{c}{\textbf{BitSqueezing}} 
		& \textbf{HGD} \\
		\cmidrule(lr){3-5} \cmidrule(lr){6-8} \cmidrule(lr){9-11}
		&                 
		& \textbf{Gaussian} & \textbf{Medium} & \textbf{Average} 
		& \textbf{Q=65} & \textbf{Q=75} & \textbf{Q=85} 
		& \textbf{4-Bits} & \textbf{5-Bits} & \textbf{6-Bits} 
		& \textbf{} \\
		\midrule
		\multirow{2}{*}{\textbf{Inc-v3}} & CGNC    
		& 34.10 & 30.21 & 26.89 
		& 28.40 & 34.09 & 40.80 
		& 36.06 & 50.10 & 52.29 
		& 52.91 \\
		& TGAF  
		& \textbf{36.59} & \textbf{45.23} & \textbf{36.25} 
		& \textbf{37.93} & \textbf{46.01} & \textbf{55.90} 
		& \textbf{51.90} & \textbf{68.36} & \textbf{70.34} 
		& \textbf{70.49} \\
		\midrule
		\multirow{2}{*}{\textbf{Res-152}} & CGNC    
		& 57.15 & 56.15 & 50.06 
		& 50.24 & 55.27 & 59.15 
		& 42.89 & 60.69 & 62.98 
		& 78.91 \\
		& TGAF 
		& \textbf{58.54} & \textbf{59.08} & \textbf{51.68} 
		& \textbf{53.46} & \textbf{57.85} & \textbf{61.22} 
		& \textbf{48.61} & \textbf{61.90} & \textbf{64.50} 
		& \textbf{83.02} \\
		\bottomrule
	\end{tabular}
    	\caption{Attack success rates (\%)  across \textbf{preprocessing technique} under VGG-16 and \textbf{HGD}.}
	\label{combined_all}
    
\end{table*}


    

\begin{table*}[!h]
\centering
  \small
  \setlength{\tabcolsep}{1mm}
\begin{tabular}{c|c|c|c|c|c|c|c|c|c|c|c}
\toprule
\textbf{Defense} & \textbf{Method} & \textbf{Inc-v3} & \textbf{Inc-v4} & \textbf{Inc-Res-v2} & \textbf{Res-152} & \textbf{DN-121} & \textbf{GoogLeNet} & \textbf{VGG-16} & \textbf{ViT-B}  & \textbf{ViF-S} & \textbf{Swin-T} \\
\hline
\multirow{2}{*}{Diffpure}
 & CGNC             & 0.43 & \textbf{0.33} & 0.21 & 0.23 & 0.40 & 0.37 & 0.21 & 0.23  & 0.24 & 0.33 \\
 & \textbf{TGAF} & \textbf{0.94} & 0.32 & \textbf{0.28} & \textbf{0.39} & \textbf{0.46} & \textbf{0.69} & \textbf{0.61} & \textbf{0.33} &  \textbf{0.51} & \textbf{0.58} \\
\hline
\multirow{2}{*}{NRP}
 & CGNC             & 12.73 & 4.25 & 2.28 & 3.10 & 5.16 & 4.51 & 4.01 & 1.61  & 4.33 & 7.43 \\
 & \textbf{TGAF} & \textbf{21.86} & \textbf{7.04} & \textbf{4.00} & \textbf{7.12} & \textbf{9.29} & \textbf{7.33} & \textbf{8.26} & \textbf{3.50} &  \textbf{9.14} & \textbf{13.56} \\
\bottomrule
\end{tabular}%

\caption{Attack success rates (\%) across \textbf{Diffpure and NRP}. The surrogate model is Inc-v3.}
\label{tab:Diffpure and NRP}
\end{table*}

\textbf{Datasets.} Following prior works ~\cite{feng2023dynamic}, the generator is trained on the ImageNet training set ~\cite{deng2009imagenet}, while the performance is evaluated on the ImageNet-NeurIPS (1k) dataset ~\cite{NIPS17}. Additional results on ImageNet Validation Set (50k) and COCO~\cite{lin2014microsoft} are provided in Appendices A and B, respectively.


\textbf{Victim Models.} We assess the effectiveness across three types of settings. 1) \textbf{Normally trained models,} including Inception-v3 (Inc‑v3) ~\cite{szegedy2016rethinking}, Inception-v4 (Inc‑v4) ~\cite{szegedy2017inception}, Inception-ResNet-v2 (Inc‑Res‑v2) ~\cite{szegedy2017inception}, ResNet-152 (Res‑152) ~\cite{he2016identity}, DenseNet-121 (DN‑121) ~\cite{huang2017densely}, GoogleNet ~\cite{szegedy2015going}, VGG-16 \cite{simonyan2014very}, ViT-B~\cite{dosovitskiy2020image}, Visformer (ViF-S)~\cite{chen2021visformer} and Swin-Tiny (Swin-T)~\cite{liu2021swin}. 2) \textbf{Robustly trained models,} which consist of adv-Inception-v3 (Inc-v3\textsubscript{ADV}) ~\cite{goodfellow2014explaining}, ens-adv-Inception-ResNet-v2 (IR-v2\textsubscript{ENS}) ~\cite{hang2020ensemble}, and several variations of resilient ResNet-50 models ~\cite{geirhos2018imagenet, hendrycks2019augmix}, including Res50\textsubscript{SIN} , Res50\textsubscript{IN}, Res50\textsubscript{FINE}~\cite{geirhos2018imagenet}), and Res50\textsubscript{AUG}~\cite{hendrycks2019augmix}). \textbf{Defense methods,} consisting of three strategie. Preprocessing techniques such as JPEG compression \cite{dziugaite2016study}, BitSqueezing \cite{xu2017feature}, and Smoothing \cite{ding2019advertorch}. Denoising methods such as the high-level representation guided denoiser (HGD) \cite{liao2018defense}. Diffusion-based methods including DiffPure~\cite{nie2022diffusion} and NRP~\cite{naseer2020self}.

\textbf{Baseline Methods.} For instance-specific attacks, we include Logit~\cite{zhao2021success}, SU~\cite{wei2023enhancing}, and Everywhere~\cite{zeng2025everywhere}. For SU and Everywhere, we use their best-performing variants: DTMI-Logit-SU and CFM-Everywhere, respectively. For instance-agnostic attacks, we compare against C-GSP~\cite{yang2022boosting} and CGNC~\cite{fang2024clip}. All baseline results are reproduced using the official code and weights.

\textbf{Implementation Details.} Following previous works ~\cite{feng2023dynamic}, we use Inc-v3 and Res-152 as surrogate models for training the generator. The reported results are averaged over 8 different target classes. The perturbation budget is set to $\epsilon = 16/255$. The generator is trained for 10 epochs with a learning rate of 2e-4 and a batch size of 16. The number of blocks, $N$, is set to 3. Specifically, the diffusion model is used only once per target class before training begins to generate the corresponding 2D tensors, which are then saved to disk. During the training and inference of TGAF, we simply load these pre-computed tensors directly.

\subsection{Evaluation on Normal Models}

We first assess the transferability of adversarial examples generated by normally trained models. Tab.~\ref{normal} demonstrates that TGAF consistently achieves state-of-the-art (SOTA) performance, exhibiting the highest black-box ASR across nearly all experimental configurations. For examples, on CNN-based models, TGAF outperforms the previous SOTA method CGNC by 16.11\%  for Inc-v3 and 5.89\% for Res-152, respectively. For Transformer-based models, the improvements are 11.17\%  for Inc-v3 and 3.84\% for Res-152.

\subsection{Evaluation on Robust Models}

\begin{table*}[!ht]
	\centering
  \small
  \setlength{\tabcolsep}{1mm}
	\begin{tabular}{c|c|c|c|c|c|c|c|c|c|c|c}
		\toprule
		\textbf{Source} & \textbf{Method} & \textbf{Inc-v3} & \textbf{Inc-v4} & \textbf{Inc-Res-v2} & \textbf{Res-152} & \textbf{DN-121} & \textbf{GoogLeNet} & \textbf{VGG-16}  & \textbf{ViT-B} &  \textbf{ViF-S} & \textbf{Swin-T}  \\
		\hline
		\multirow{2}{*}{\textbf{Inc-v3}}& TGAF-C & 96.23* & 64.21 & 51.60 & 50.11 & 70.18 & 61.36 & 60.55 & 27.80 & 50.96 & 25.17 \\
		&TGAF & \textbf{98.15*} & \textbf{72.49} & \textbf{63.20} & \textbf{61.94} & \textbf{78.30} & \textbf{66.86} & \textbf{70.64}  & \textbf{33.03} &  \textbf{63.27} & \textbf{42.61}  \\
		\hline
		\multirow{2}{*}{\textbf{Res-152}} & TGAF-N & 56.28 & 54.95 & 40.60 & 95.26* & 83.28 & 62.95 & 61.84 & 33.53  & 53.28 &31.76   \\
		& TGAF & \textbf{62.60} & \textbf{62.44} &\textbf{44.03} & \textbf{97.79*} & \textbf{87.90} & \textbf{63.54} & \textbf{65.20} & \textbf{39.64} &  \textbf{63.36} & \textbf{42.84}\\
		\bottomrule
	\end{tabular}

    \caption{\textbf{Ablation study on masking strategies.} TGAF-C represents the substitution of our masking strategy with the CGNC masking strategy, while TGAF-N represents the removal of our masking strategy.}
	\label{tab:combined_masking}

\end{table*}

\begin{table*}[ht]
\centering
  \small
  \setlength{\tabcolsep}{1mm}
\begin{tabular}{c|c|c|c|c|c|c|c|c|c|c}
\toprule
\textbf{Method} & \textbf{Inc-v3} & \textbf{Inc-v4} & \textbf{Inc-Res-v2} & \textbf{Res-152} & \textbf{DN-121} & \textbf{GoogleNet} & \textbf{VGG-16} & \textbf{ViT-B} &  \textbf{ViF-S} & \textbf{Swin-T} \\
\hline
TGAF      & \textbf{62.60} & \textbf{62.44} & \textbf{44.02} & \textbf{97.79*} & \textbf{87.90} & \textbf{63.54} & \textbf{65.20} & 39.64 &  63.36 & 42.84 \\
TGAF-Conv & 59.65 & 60.24 & 38.80 & 97.49* & 87.76 & 59.38 & 63.78 & \textbf{40.80} &  \textbf{65.99} & \textbf{43.09} \\
TGAF-CA   & 47.34 & 46.75 & 32.18 & 93.10* & 76.96 & 53.15 & 56.56 & 28.95  & 45.19 & 27.10 \\
\bottomrule
\end{tabular}%
\caption{\textbf{Ablation study on fusion strategy.} TGAF-Conv represents the TGAF model without the CbF module. TGAF-CA represents the TGAF model without the TbF module. The surrogate model is Res-152.}
\label{tab:fusion_comparison}

\end{table*}

\begin{table*}[!h]
\centering
  \small
  \setlength{\tabcolsep}{1mm}
\begin{tabular}{c|c|c|c|c|c|c|c|c|c|c}
\toprule
\textbf{Block (N)} & \textbf{Inc-v3} & \textbf{Inc-v4} & \textbf{Inc-Res-v2} & \textbf{Res-152} & \textbf{DN-121} & \textbf{GoogleNet} & \textbf{VGG-16} & \textbf{ViT-B}  & \textbf{ViF-S} & \textbf{Swin-T} \\
\hline
2          & 55.70            & 52.64            & 35.66               & 97.25*            & 84.34            & 56.25             & 57.01           & 32.96                      & 58.36           & 40.96            \\
\textbf{3} & \textbf{62.60}   & \textbf{62.44}   & \textbf{44.02}      & \textbf{97.79*}   & \textbf{87.90}   & 63.54    & 65.20  & \textbf{39.64}  & \textbf{63.36}  & \textbf{42.84}   \\
4          & 60.31            & 59.79            & 40.73               & 97.68*            & 86.91            & \textbf{64.95}    & \textbf{68.09}  & 35.71                    & 60.20           & 41.95            \\
\bottomrule
\end{tabular}%
\caption{\textbf{Analysis of block number sensitivity.} The surrogate model is Res-152.}

\label{tab:block_Sensitivity}
\end{table*}

Building upon the findings obtained with normally trained models, we proceed to evaluate the performance of our method on robustly trained models. Tab.~\ref{robust}
shows that our method also achieves SOTA performance. Specifically, when using Inc‑v3 and Res-152 as surrogate models, our approach delivers average ASR improvements of 16.92\% and 5.66\% over CGNC, and exceeds other methods by more than 30\% and 10\%, respectively.

\subsection{Evaluation on Defense Methods}

To further evaluate the robustness, we evaluate  TGAF against multiple defense mechanisms. Tabs.~\ref{combined_all} and~\ref{tab:Diffpure and NRP} show that TGAF outperforms CGNC across nearly all tested defense mechanisms.
It's important to note that both DiffPure and NRP significantly reduce the targeted attack success rate. This is an expected outcome, as these defenses are designed to substantially alter input information (or its internal representations) to "cleanse" adversarial perturbations. This purification process, however, is inherently lossy. While it effectively removes the perturbation, it can inadvertently corrupt the original semantic features crucial for correct classification. Therefore, a critical point must be emphasized: a low ASR does not necessarily mean the model recovered the correct classification. In many cases, the purified image, while not misclassified as the attacker's intended target, is instead misclassified as a different, incorrect class due to this information loss. Despite these robust defenses, TGAF  consistently and substantially outperforms the CGNC baseline. For example, when tested against NRP on Swin-T, TGAF achieves a ASR of 13.56\%, which is nearly double that of CGNC's 7.43\%.

	\subsection{Evaluation on Adversarial Sample Quality}

   Tab.~\ref{tab:Quality} shows that  TGAF exhibits only marginal differences in SSIM, LPIPS, and FID metrics compared to CGNC. For instance, the LPIPS differs by merely 0.013. Simultaneously, TGAF achieves a slightly better PSNR score. We contend that this negligible difference in perceptual quality represents a highly successful trade-off for a substantial improvement in attack transferability.

    \begin{table}[ht]
\centering
  \small
  \setlength{\tabcolsep}{1mm}
\begin{tabular}{ccccc}
\toprule
\textbf{Method} & \textbf{PSNR ($\uparrow$)} & \textbf{SSIM ($\uparrow$)} & \textbf{LPIPS ($\downarrow$)} & \textbf{FID ($\downarrow$)} \\
\midrule
CGNC             & 27.986                     & \textbf{0.814}             & \textbf{0.186}               & \textbf{244.633}             \\
\textbf{TGAF} & \textbf{27.994}            & 0.801                      & 0.199                        & 259.945                      \\
\bottomrule
\end{tabular}
\caption{\textbf{Adversarial sample quality.} (Res-152 surrogate)}
\label{tab:Quality}
\end{table}

    	\subsection{Ablation Study}
	\label{subsec:AblationStudy}

To thoroughly evaluate the effectiveness of TGAF, we conduct comprehensive ablation studies on masking strategies, fusion mechanisms, and block number sensitivity. Tab.~\ref{tab:combined_masking} analyzes the effects of the masking strategy. Results demonstrate that when Inc-v3 serves as the surrogate model, our method achieves an average ASR improvement of  9.24\% for CNN-based models  and 11.66\% for Transformer-based models. Furthermore, compared to TGAF-N with Res-152 as the surrogate model, our masking strategy leads to a  enhancement of 4.31\% for CNN-based models and 9.10\% for Transformer-based models. Tab.~\ref{tab:fusion_comparison} details the performance on fusion mechanisms. It shows that removing the TbF led to a significant performance degradation across all black-box models.
For the CbF, removing  it resulted in a slight decrease in ASR against most CNN models (an average drop of about 2.68\%). This indicates that the CbF module positively contributes to attacking CNN models by learning local feature interactions. Interestingly, we observed that the performance of TGAF-Conv slightly improved when attacking Transformer architectures. We hypothesize that this is because CbF learns features that might slightly overfit to CNN-style architectures, thereby hindering transferability to Transformer models. Tab.~\ref{tab:block_Sensitivity} investigates the sensitivity to the number of blocks ($N$). The results show that setting $N=3$ consistently outperforms $N=2$ across all 10 target models. When comparing $N=3$ and $N=4$, we observe that while $N=4$ performs marginally better on certain classic CNNs, $N=3$ exhibits stronger overall performance on the majority of models. Crucially, the attack performance remains at a very high level for both $N=3$ and $N=4$, without any sharp drop in ASR. This robust performance indicates that the success of our method stems from its core design principles and a well-balanced architecture, rather than a fragile dependency on a highly fine-tuned hyperparameter.

\section{Conclusion}

In this paper, we identify that the quality and quantity of the implanted target semantic features are the key factors influencing the transferability of adversarial attacks. To address these challenges, we propose TGAF, a novel method that leverages the abilities of diffusion models to encode target labels as 2D tensors, improving the quality of target semantic information. Furthermore, we introduce a masking strategy during the training phase to ensure that the portion of the generated noise retains complete target semantic features, thus enhancing the quantity.
Our extensive experiments demonstrate that TGAF not only surpasses existing SOTA methods on normally trained models but also shows robust performance against a variety of defense strategies. We hope that our proposed method serves as a reliable tool for evaluating model robustness under black-box attack settings and provides new insight for further research on vulnerability and robustness in adversarial scenarios.

\section*{Acknowledgments}
This work was supported by the National Science and Technology Innovation 2030 - Major Project (Grant No. 2022ZD0208800), and NSFC General Program (Grant No. 62176215).


\bibliography{main}

\appendix

\section*{Appendix}

\section{Evaluation on ImageNet Validation Set}
For a more comprehensive analysis, we compare the multi-target performance of CGNC and TGAF on the whole ImageNet~\cite{deng2009imagenet} validation dataset (50k), and the results are summarized in Tab.~\ref{tab:attack_rates}. Evidently, TGAF produces consistently stronger attack rates against a variety of normally trained models. In particular, when Inc-v3 is used as the source, TGAF achieves a 71.69\% success rate on Inc-v4, outperforming CGNC by over 13\%. A similar trend is observed with Res-152 as the surrogate. In general, TGAF surpasses CGNC across all target models on Inc-v3 and Res-152 with average improvement of 14.96\% and 4.96\% respectively, while maintaining near-maximum white-box rates. These findings confirm the superior transferability of TGAF in multi-target adversarial attacks.

\begin{table*}[htbp]
\centering
\begin{tabular}{c|c|c|c|c|c|c|c|c}
\toprule
\textbf{Source} & \textbf{Method} & \textbf{Inc-v3} & \textbf{Inc-v4} & \textbf{Inc-Res-v2} & \textbf{Res-152} & \textbf{DN-121} & \textbf{GoogleNet} & \textbf{VGG-16} \\
\hline
\multirow{2}{*}{\textbf{Inc-v3}} & CGNC
    & 96.60* & 57.85 & 46.88 & 44.15 & 65.91 & 53.42 & 56.29 \\
& TGAF & \textbf{98.45*} & \textbf{71.69} & \textbf{61.47} & \textbf{61.67} & \textbf{79.67} & \textbf{67.19} & \textbf{72.58} \\
\hline
\multirow{2}{*}{\textbf{Res-152}} & CGNC
    & 56.03 & 50.40 & 32.29 & 96.45* & 86.69 & 63.86 & 63.94 \\
 & TGAF
    & \textbf{63.63} & \textbf{61.30} & \textbf{42.03} & \textbf{97.99*} & \textbf{87.91} & \textbf{64.14} & \textbf{63.99} \\
\bottomrule
\end{tabular}
\caption{Attack success rates (\%) for multi-target attacks against normally trained models on ImageNet validation set (50k). The results are averaged over 8 different target classes. * represents white-box attacks.}
\label{tab:attack_rates}
\end{table*}

\section{Evaluation on COCO dataset}

For evaluation on the COCO dataset~\cite{lin2014microsoft}, we randomly sampled 1000 images and conducted a direct comparison with CGNC. The detailed results are presented in Tab.~\ref{tab:coco}. Our TGAF method demonstrates superior performance: on CNN-based models, it surpasses CGNC by 14.84\% for Inc-v3 and 5.29\% for Res-152, respectively. For Transformer-based models, the improvements are 14.48\% for Inc-v3 and 3.95\% for Res-152.


\begin{table*}[!ht]
  \centering
  
  \resizebox{\textwidth}{!}{%
  \begin{tabular}{c|c|c|c|c|c|c|c|c|c|c|c}
    \toprule
    \textbf{Source} & \textbf{Method} & \textbf{Inc-v3} & \textbf{Inc-v4} & \textbf{Inc-Res-v2} & \textbf{Res-152} & \textbf{DN-121} & \textbf{GoogLeNet} & \textbf{VGG-16} & \textbf{ViT-B}  & \textbf{ViF-S} & \textbf{Swin-T} \\
    \hline
    \multirow{2}{*}{\textbf{Inc-v3}}
     & CGNC       & 99.01* & 73.36 & 60.75 & 52.01 & 73.79 & 60.74 & 67.36 & 28.07  & 62.47 & 32.25 \\
     & \textbf{TGAF}    & \textbf{99.73*} & \textbf{84.86} & \textbf{76.86} & \textbf{71.20} & \textbf{86.67} & \textbf{75.77} & \textbf{80.47} & \textbf{40.53}  & \textbf{74.44} & \textbf{51.31} \\
    \hline
    \multirow{2}{*}{\textbf{Res-152}}
     & CGNC      
     & 61.28 & 60.40 & 41.23 & 99.11* & 92.26 & 71.30 & 75.73 & 40.38 &  70.02 & 54.29 \\
     
     & \textbf{TGAF}     & \textbf{69.72} & \textbf{71.28} & \textbf{51.39} & \textbf{99.74*} & \textbf{93.38} & \textbf{71.92} & \textbf{76.23} & \textbf{43.80}  & \textbf{75.03} & \textbf{57.72} \\
    \bottomrule
  \end{tabular}%
  }
 \caption{Attack success rates (\%) for multi-target attacks against normally trained models on COCO. The results are averaged over 8 different target classes. * represents white-box attacks.}
  \label{tab:coco}
\end{table*}

\section{Extended Evaluations on Defense Methods}
In this section, we present additional experiments on the defense methods. The generator is trained on the ImageNet training set ~\cite{deng2009imagenet}, while the adversarial attack performance is evaluated on the ImageNet NeurIPS (1k) dataset ~\cite{NIPS17}. We evaluate our approach under three defense scenarios, including Smoothing, JPEG Compression, and BitSqueezing  with target models (Inc-v3, Inc-v4, Inc-Res-v2, Res-152, DN-121 and GoogleNet).
The comprehensive results are summarized in Tab.~\ref{smoothing} through Tab.~\ref{bit}.

Specifically, Tab.~\ref{smoothing} presents the attack success rates (\%) for multi-target attacks under the Smoothing Defense. Our method consistently outperforms CGNC under all three smoothing strategies (Gaussian, Median, and Average). On Inc-v4, the rate goes from 60.58\% to 73.86\% (an improvement of 13.3\%), and on Inc-Res-v2 from 43.97\% to 57.33\% (approximately 13.4\%). In the case of Median smoothing, improvements of around 3.7\%, 12.1\%, and 14.4\% were observed on Inc-v3, Inc-v4, and Inc-Res-v2 respectively. Across the evaluated target models, TGAF delivers an average improvement of approximately 10\% in attack success rates relative to CGNC.


Tab.~\ref{jpeg} summarizes the performance of our proposed method under the JPEG Compression Defense. Utilizing Inc-v3 as the surrogate model, our approach demonstrates significant gains in attack success rates compared to the baseline method. For instance, at a quality factor of $Q=65$, the success rate on Inc-Res-v2 improves from $38.90\%$ (CGNC) to $52.95\%$ (TGAF), reflecting an increase of over $14\%$. Similarly, the success rate on Res-152 rises from $24.66\%$ to $38.89\%$ (approximately $14.2\%$), while DenseNet-121 sees an increase from $42.01\%$ to $53.66\%$ (around $11.7\%$). Comparable improvements are observed at quality factors $Q=75$ and $Q=85$. On average, across all quality factors and target models, our method achieves approximately $12\%$ higher success rates compared to the baseline approach. These results highlight the robust effectiveness of the TGAF framework in scenarios where image quality is intentionally degraded, further validating its capability in challenging defense conditions.

In Tab.~\ref{bit}, we report the attack success rates under BitSqueezing Defense. The evaluation is performed at three different quantization levels (4, 5, and 6 bits). Our experimental results indicate that TGAF boosts the attack success rates by about 13\% on average over CGNC across the evaluated target models. This result underscores the robustness of our technique even under extreme defense conditions.

\begin{table*}[!ht]
    \centering
    
    \resizebox{\textwidth}{!}{%
    \begin{tabular}{c|c|c|c|c|c|c|c|c|c}
    \toprule
    \textbf{Source} & \textbf{Method} & \textbf{Smoothing} 
      & \textbf{Inc-v3} & \textbf{Inc-v4} & \textbf{Inc-Res-v2} 
      & \textbf{Res-152} & \textbf{DN-121} & \textbf{GoogleNet} & \textbf{VGG-16} \\
    \hline
    \multirow{6}{*}{\textbf{Inc-v3}} 
      & \multirow{3}{*}{CGNC} 
        & G & 93.85* & 60.58 & 43.97 & 25.31 & 38.80 & 26.71 & 34.10 \\
      &  & M & 94.00* & 61.18 & 41.94 & 23.49 & 44.50 & 28.69 & 30.21 \\
      &  & A & 92.28* & 57.39 & 40.77 & 19.15 & 31.97 & 21.15 & 26.89 \\
    \cline{2-10}
      & \multirow{3}{*}{\textbf{TGAF}} 
        & G & \textbf{97.75*} & \textbf{73.86} & \textbf{57.33} & \textbf{35.65} & \textbf{47.50} & \textbf{35.09} & \textbf{36.59} \\
      &  & M & \textbf{97.65*} & \textbf{73.29} & \textbf{56.29} & \textbf{39.09} & \textbf{57.39} & \textbf{37.05} & \textbf{45.23} \\
      &  & A & \textbf{97.24*} & \textbf{70.91} & \textbf{52.17} & \textbf{25.95} & \textbf{37.75} & \textbf{26.04} & \textbf{36.25} \\
    \hline
    \multirow{6}{*}{\textbf{Res-152}} 
      & \multirow{3}{*}{CGNC} 
        & G & 44.69 & 48.59 & 28.66 & 92.78* & 74.95 & 44.24 & 57.15 \\
      &  & M & 49.38 & 51.70 & 30.64 & 93.73* & 79.89 & 46.81 & 56.15 \\
      &  & A & 41.15 & 45.98 & 26.71 & 90.20* & 69.23 & 37.09 & 50.06 \\
    \cline{2-10}
      & \multirow{3}{*}{\textbf{TGAF}} 
        & G & \textbf{52.25} & \textbf{57.66} & \textbf{37.06} & \textbf{96.26*} & \textbf{75.61} & \textbf{45.96} & \textbf{58.54} \\
      &  & M & \textbf{58.98} & \textbf{61.30} & \textbf{39.69} & \textbf{96.29*} & \textbf{81.67} & \textbf{48.04} & \textbf{59.08} \\
      &  & A & \textbf{47.95} & \textbf{54.75} & \textbf{34.16} & \textbf{94.62*} & \textbf{69.40} & \textbf{37.41} & \textbf{51.68} \\
    \bottomrule
    \end{tabular}%
    }
    \caption{Attack Success Rates (\%) for multi-target attacks under Smoothing Defense. The smoothing methods are Gaussian (G), Median (M), and Average (A). The results are averaged over 8 different target classes. * represents white-box attacks.}
    \label{smoothing}
\end{table*}

\begin{table*}[!ht]
    \centering
   
    \resizebox{\textwidth}{!}{%
    \begin{tabular}{c|c|c|c|c|c|c|c|c|c}
    \toprule
    \textbf{Source} & \textbf{Method} & \textbf{Quality} 
      & \textbf{Inc-v3} & \textbf{Inc-v4} & \textbf{Inc-Res-v2} 
      & \textbf{Res-152} & \textbf{Dense121} & \textbf{GoogleNet} & \textbf{VGG-16} \\
    \hline
    \multirow{6}{*}{\textbf{Inc-v3}} 
      & \multirow{3}{*}{CGNC} 
        & Q=65 & 91.14* & 47.45 & 38.90 & 24.66 & 42.01 & 30.34 & 28.40 \\
      &  & Q=75 & 92.65* & 50.94 & 41.59 & 29.73 & 48.79 & 35.78 & 34.09 \\
      &  & Q=85 & 94.06* & 53.23 & 44.70 & 36.33 & 55.58 & 42.10 & 40.80 \\
    \cline{2-10}
      & \multirow{3}{*}{\textbf{TGAF}} 
        & Q=65 & \textbf{96.28*} & \textbf{60.65} & \textbf{52.95} & \textbf{38.89} & \textbf{53.66} & \textbf{38.95} & \textbf{37.93} \\
      &  & Q=75 & \textbf{96.88*} & \textbf{62.84} & \textbf{56.44} & \textbf{45.09} & \textbf{61.31} & \textbf{46.23} & \textbf{46.01} \\
      &  & Q=85 & \textbf{97.39*} & \textbf{65.75} & \textbf{59.73} & \textbf{52.63} & \textbf{69.19} & \textbf{55.09} & \textbf{55.90} \\
    \hline
    \multirow{6}{*}{\textbf{Res-152}} 
      & \multirow{3}{*}{CGNC} 
        & Q=65 & 42.26 & 44.06 & 26.70 & 88.26* & 75.50 & 43.31 & 50.24 \\
      &  & Q=75 & 45.61 & 46.23 & 29.38 & 90.71* & 78.85 & 47.63 & 55.27 \\
      &  & Q=85 & 48.66 & 48.56 & 31.84 & 92.66* & 81.81 & 52.85 & 59.15 \\
    \cline{2-10}
      & \multirow{3}{*}{\textbf{TGAF}} 
        & Q=65 & \textbf{53.59} & \textbf{54.89} & \textbf{36.96} & \textbf{92.71*} & \textbf{77.85} & \textbf{46.96} & \textbf{53.46} \\
      &  & Q=75 & \textbf{56.01} & \textbf{57.25} & \textbf{39.76} & \textbf{93.85*} & \textbf{81.09} & \textbf{51.04} & \textbf{57.85} \\
      &  & Q=85 & \textbf{59.26} & \textbf{59.31} & \textbf{42.50} & \textbf{95.21*} & \textbf{83.82} & \textbf{55.41} & \textbf{61.22} \\
    \bottomrule
    \end{tabular}%
    }
     \caption{Attack Success Rates (\%) for multi-target attacks under JPEG Compression Defense. The quality factors are Q=65, Q=75, and Q=85. The results are averaged over 8 different target classes. * represents white-box attacks.}
    \label{jpeg}
\end{table*}

\begin{table*}[!ht]
  \centering
  
  \resizebox{\textwidth}{!}{%
  \begin{tabular}{c|c|c|c|c|c|c|c|c|c}
    \toprule
    \textbf{Source} & \textbf{Method} & \textbf{Bits} 
      & \textbf{Inc-v3} & \textbf{Inc-v4} & \textbf{Inc-Res-v2} 
      & \textbf{Res-152} & \textbf{DN-121} & \textbf{GoogleNet} & \textbf{VGG-16} \\
    \hline
    \multirow{6}{*}{\textbf{Inc-v3}} 
      & \multirow{3}{*}{CGNC} 
        & 4 & 95.05* & 48.20 & 39.38 & 26.88 & 48.90 & 40.51 & 36.06 \\
      &  & 5 & 95.89* & 56.69 & 46.03 & 39.90 & 60.20 & 49.20 & 50.10 \\
      &  & 6 & 96.00* & 58.74 & 47.35 & 42.04 & 62.36 & 50.94 & 52.29 \\
    \cline{2-10}
      & \multirow{3}{*}{\textbf{TGAF}} 
        & 4 & \textbf{97.93*} & \textbf{62.14} & \textbf{53.95} & \textbf{43.35} & \textbf{65.33} & \textbf{54.18} & \textbf{51.90} \\
      &  & 5 & \textbf{98.11*} & \textbf{70.36} & \textbf{61.00} & \textbf{58.26} & \textbf{76.61} & \textbf{64.59} & \textbf{68.36} \\
      &  & 6 & \textbf{98.14*} & \textbf{71.80} & \textbf{62.55} & \textbf{61.29} & \textbf{78.19} & \textbf{66.40} & \textbf{70.34} \\
    \hline
    \multirow{6}{*}{\textbf{Res-152}} 
      & \multirow{3}{*}{CGNC} 
        & 4 & 43.24  & 31.38  & 21.08  & 92.71* & 78.26  & 50.84  & 42.89 \\
      &  & 5 & 52.80  & 48.14  & 31.73  & 95.55* & 84.14  & 60.03  & 60.69 \\
      &  & 6 & 53.75  & 51.11  & 33.66  & 95.76* & 85.20  & 61.99  & 62.98 \\
    \cline{2-10}
      & \multirow{3}{*}{\textbf{TGAF}} 
        & 4 & \textbf{54.30} & \textbf{42.29} & \textbf{30.10} & \textbf{96.76*} & \textbf{82.03} & \textbf{53.26} & \textbf{48.61} \\
      &  & 5 & \textbf{61.75} & \textbf{58.31} & \textbf{40.89} & \textbf{97.69*} & \textbf{86.79} & \textbf{60.93} & \textbf{61.90} \\
      &  & 6 & \textbf{62.83} & \textbf{61.64} & \textbf{43.24} & \textbf{97.79*} & \textbf{87.71} & \textbf{62.86} & \textbf{64.50} \\
    \bottomrule
  \end{tabular}%
  }
  \caption{Attack Success Rates (\%) for multi-target attacks under BitSqueezing Defense. The varying quantization levels are 4, 5, and 6 bits. The results are averaged over 8 different target classes. * represents white-box attacks.}
  \label{bit}
\end{table*}

\section{Additional Ablation Studies}

In this section, we present ablation experiments focusing on the 2D tensors within our modules. Specifically, as shown in Tab.~\ref{tab:appendixabalation}, C-TGAF refers to replacing our 2D tensors with CGNC 1D tensors generated by CLIP. We evaluated the differences when using Inc-v3 and Res-152 as surrogate models. The results demonstrate that the full TGAF consistently outperforms C-TGAF, highlighting the effectiveness of our approach.

\begin{table*}[!ht]
    \centering
    \resizebox{0.9\textwidth}{!}{
	\begin{tabular}{c|c|c|c|c|c|c|c|c|c|c|c}
		\toprule
		\textbf{Source} & \textbf{Method} & \textbf{Inc-v3} & \textbf{Inc-v4} & \textbf{Inc-Res-v2} & \textbf{Res-152} & \textbf{DN-121} & \textbf{GoogLeNet} & \textbf{VGG-16}  & \textbf{ViT-B}  & \textbf{ViF-S} & \textbf{Swin-T}  \\
		\hline
		\multirow{2}{*}{\textbf{Inc-v3}}
		&C-TGAF & 96.95* & 65.21 & 55.18 & 56.68 & 71.79 & 58.82 & 68.22 &31.51  & \textbf{64.51 }&40.74   \\
		&TGAF & \textbf{98.15*} & \textbf{72.49} & \textbf{63.20} & \textbf{61.94} & \textbf{78.30} & \textbf{66.86} & \textbf{70.64}  & \textbf{33.03}  & 63.27 & \textbf{42.61}  \\
		\hline
		\multirow{2}{*}{\textbf{Res-152}} 
		&C-TGAF & 56.44 & 56.57 & 36.35 & 96.31* & 86.15 & 61.94 & 62.24 & 39.35 & \textbf{64.61} &40.08   \\
		& TGAF & \textbf{62.60} & \textbf{62.44} &\textbf{44.03} & \textbf{97.79*} & \textbf{87.90} & \textbf{63.54} & \textbf{65.20} & \textbf{39.64}  & 63.36 & \textbf{42.84} \\
		\bottomrule
	\end{tabular}
    }
     \caption{Ablation study of TGAF on 2D Tensor: C-TGAF denotes replacing  2D tensors with 1D tensors generated by CLIP.}
    \label{tab:appendixabalation}
\end{table*}





\section{Comparison with more Methods}
In this section, we present a comparative analysis of our TGAF method against SASD-WS~\cite{wu2024improving} and GAKer~\cite{sun2024any}. For fair comparison, all experiments used the pre-trained model weights provided by the original authors.

As detailed in Tab.~\ref{tab:SASD}, our TGAF consistently outperforms SASD-WS across all victim models when Inc-v3 is employed as the surrogate model. Nevertheless, SASD-WS demonstrates exceptional performance when Res-152 is the surrogate. While our method did not exceed SASD-WS in this particular setup, TGAF still yielded commendable attack success rates, underscoring its robustness as a versatile attack framework. A crucial consideration is the inherent trade-off between attack effectiveness and computational demands. As highlighted in literature~\cite{zeng2025everywhere}, advanced iterative techniques can sometimes achieve superior success rates compared to generation-based approaches like ours. However, these iterative methods suffer from substantial computational overhead, necessitating N optimization iterations per image, which is time-consuming and less feasible for large-scale applications. In contrast, our generation-based TGAF generates an attack with just a single forward pass during inference.

Furthermore, Tab.~\ref{tab:GAKer} illustrates TGAF's notable advantage over GAKer, also a generation-based method, consistently achieving superior performance across all victim models.

\begin{table*}[!h]
\centering
\resizebox{\textwidth}{!}{%
\begin{tabular}{c|c|c|c|c|c|c|c|c|c|c|c}
\toprule
\textbf{Source} & \textbf{Method} & \textbf{Inc-v3} & \textbf{Inc-v4} & \textbf{Inc-Res-v2} & \textbf{Res-152} & \textbf{DN-121} & \textbf{GoogleNet} & \textbf{VGG-16} & \textbf{ViT-B} & \textbf{ViF-S} & \textbf{Swin-T} \\
\hline
\multirow{2}{*}{\textbf{Inc-v3}}
 & SASD\_WS        & 91.75*  & 33.16  & 28.30  & 25.79  & 48.63  & 48.88  & 45.21  & 6.49   & 18.24  & 9.13   \\
 & \textbf{TGAF}& \textbf{98.15*} & \textbf{72.49} & \textbf{63.20} & \textbf{61.94} & \textbf{78.30} & \textbf{66.86} & \textbf{70.64} & \textbf{33.03} & \textbf{25.10} & \textbf{42.61} \\
\hline
\multirow{2}{*}{\textbf{Res-152}}
 & SASD\_WS        & \textbf{80.61}  & \textbf{66.89}  & \textbf{65.95}  & \textbf{99.15*}  & 9\textbf{8.74}  & \textbf{94.41}  & \textbf{74.14}  & \textbf{45.68}  & \textbf{79.38}  & \textbf{57.30}  \\
 & \textbf{TGAF}& 62.60 & 62.44 & 44.02 & 97.79* & 87.90 & 63.54 & 65.20 & 39.64 & 63.36 & 42.84 \\
\bottomrule
\end{tabular}%
}
\caption{Attack success rates (\%) of SASD\_WS and TGAF  on ImageNet NeurIPS validation set. The results are averaged over 8 different target classes. * represents white-box attacks.}
\label{tab:SASD}
\end{table*}

\begin{table*}[!h]
\centering
\resizebox{\textwidth}{!}{%
\begin{tabular}{c|c|c|c|c|c|c|c|c|c|c|c}
\toprule
\textbf{Method}    & \textbf{Inc-v3} & \textbf{Inc-v4} & \textbf{Inc-Res-v2} & \textbf{Res-152} & \textbf{DN-121} & \textbf{GoogleNet} & \textbf{VGG-16} & \textbf{ViT-B} & \textbf{ViF-S} & \textbf{Swin-T} \\
\hline
GAKer              & 24.18           & 22.79           & 13.09               & 49.15            & 52.66           & 25.81             & 47.46           & 10.74          & 30.97          & 18.08           \\
\textbf{TGAF}   & \textbf{55.68}  & \textbf{54.81}  & \textbf{30.51}      & \textbf{85.55}   & \textbf{90.25}  & \textbf{61.70}    & \textbf{72.41}  & \textbf{34.69} & \textbf{64.64} & \textbf{42.84}  \\
\bottomrule
\end{tabular}%
}
\caption{Attack success rates (\%) of GAKer and TGAF  on ImageNet NeurIPS validation set, with Res-50 as the Surrogate Model. The results are averaged over 8 different target classes. }
\label{tab:GAKer}
\end{table*}

\begin{table*}[htbp]
\centering
\begin{tabular}{l r r r} 
\toprule
\textbf{Metric} & \textbf{TGAF} & \textbf{CGNC} & \textbf{Relative Overhead} \\
\midrule
Training Time (per epoch) & 28.5 h & 26.7 h & $\sim$1.07x \\
Peak Training GPU Memory & 39.32 GB & 37.12 GB & $\sim$1.06x \\
\bottomrule
\end{tabular}
\caption{
    Comparison of computational costs on a single NVIDIA A100 GPU. TGAF introduces a minor computational overhead, demonstrating its practical feasibility for the significant improvement in attack transferability.
}
\label{tab:computational_cost}
\end{table*}



\section{Feature Quality Evaluation}

To quantitatively evaluate the quality of features in the generated perturbations, we conducted experiments on 1000 images from the ImageNet dataset. Specifically, we computed the cosine similarity between the features extracted from the generated adversarial perturbations ($\Delta x$) and the average features of real images from the target class. The cosine similarity is calculated as follows:

\begin{equation}
\label{eq:feature_quality}
\text{Feature Quality} = \frac{f(\Delta x) \cdot f_{\text{avg}}(x_{\text{real}})}{||f(\Delta x)|| \cdot ||f_{\text{avg}}(x_{\text{real}})||}.
\end{equation}

Here, $f$ denotes a pre-trained model such as ResNet-50. The term $\Delta x$ represents the generated adversarial perturbation, which is the difference between an adversarial example $x^{adv}$ and its corresponding original clean image $x$. Before being passed through $f$, $\Delta x$ is first normalized to a $[0, 1]$ range using min-max scaling and then standardized using ImageNet's mean and standard deviation. The term $f_{\text{avg}}(x_{\text{real}})$ signifies the pre-computed average feature vector for real images belonging to the target class. This average feature vector is obtained by applying the same feature extractor $f$ to a collection of real images from the target class and subsequently averaging their extracted features.

\section{Feature Quantity Evaluation}

To quantitatively evaluate the feature quantity on adversarial examples, we conducted experiments on 1000 images from the ImageNet dataset. Specifically, we computed the ratio of the activated area within the Grad-CAM attention map to the total area of the map, for a given target class. The attention area ratio is calculated as follows:

\begin{equation}
\label{eq:feature_quantity}
\text{Attention Area Ratio} = \frac{\sum \mathbb{I}(M_{CAM} > \tau)}{\text{Total Pixels in } M_{CAM}}.
\end{equation}

Here, $M_{CAM}$ represents the Grad-CAM activation map for the adversarial example ($x^{adv}$) with respect to the target class.  $\mathbb{I}(\cdot)$ is the indicator function, and $\tau$ is a predefined threshold. The sum $\sum \mathbb{I}(M_{CAM} > \tau)$ counts the number of pixels in $M_{CAM}$ whose normalized activation values exceed $\tau$. The Total Pixels in $M_{CAM}$ refers to the total number of pixels in the activation map.

\section{Display of Prompts}
\begin{figure*}[hbtp]
\centering
\includegraphics[width=0.9\textwidth]{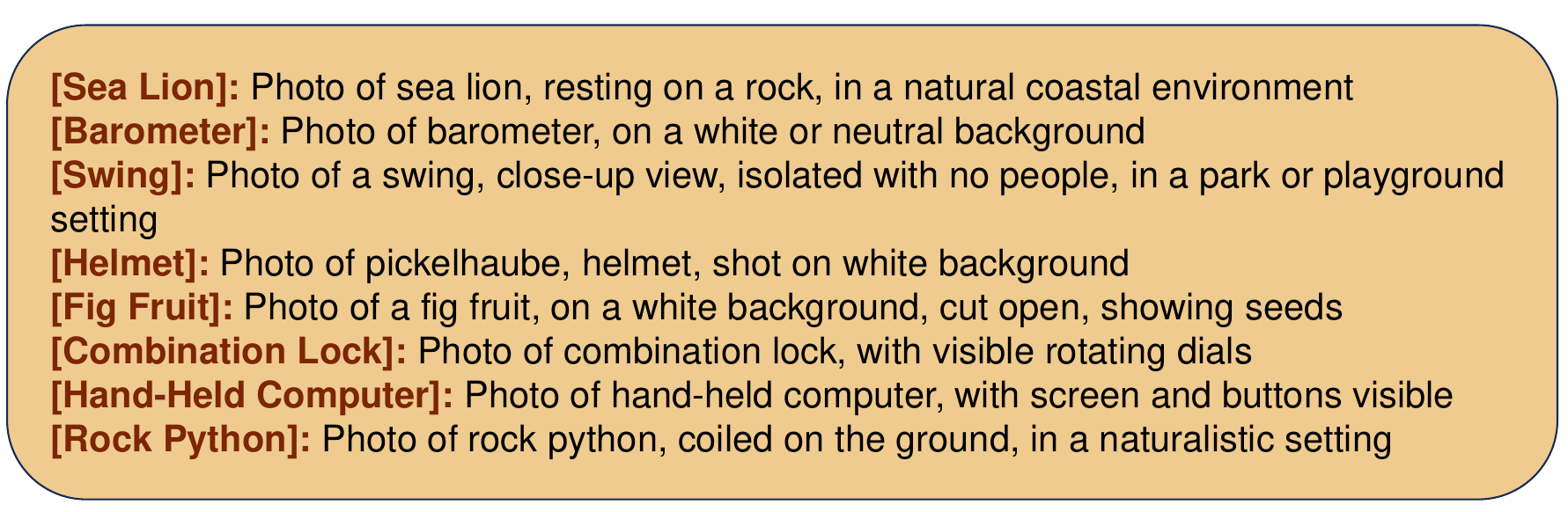}
\caption{
\textbf{Prompts used for latent tensors utilized for generating the target image.}
}
\label{fig:app2}
\end{figure*}

This section details the prompts employed to derive the 2D-Tensors. As illustrated in Fig.~\ref{fig:app2}, each prompt corresponds to a particular target class and is carefully designed. 

\begin{figure*}[!htbp]
\centering
\includegraphics[width=0.9\textwidth]{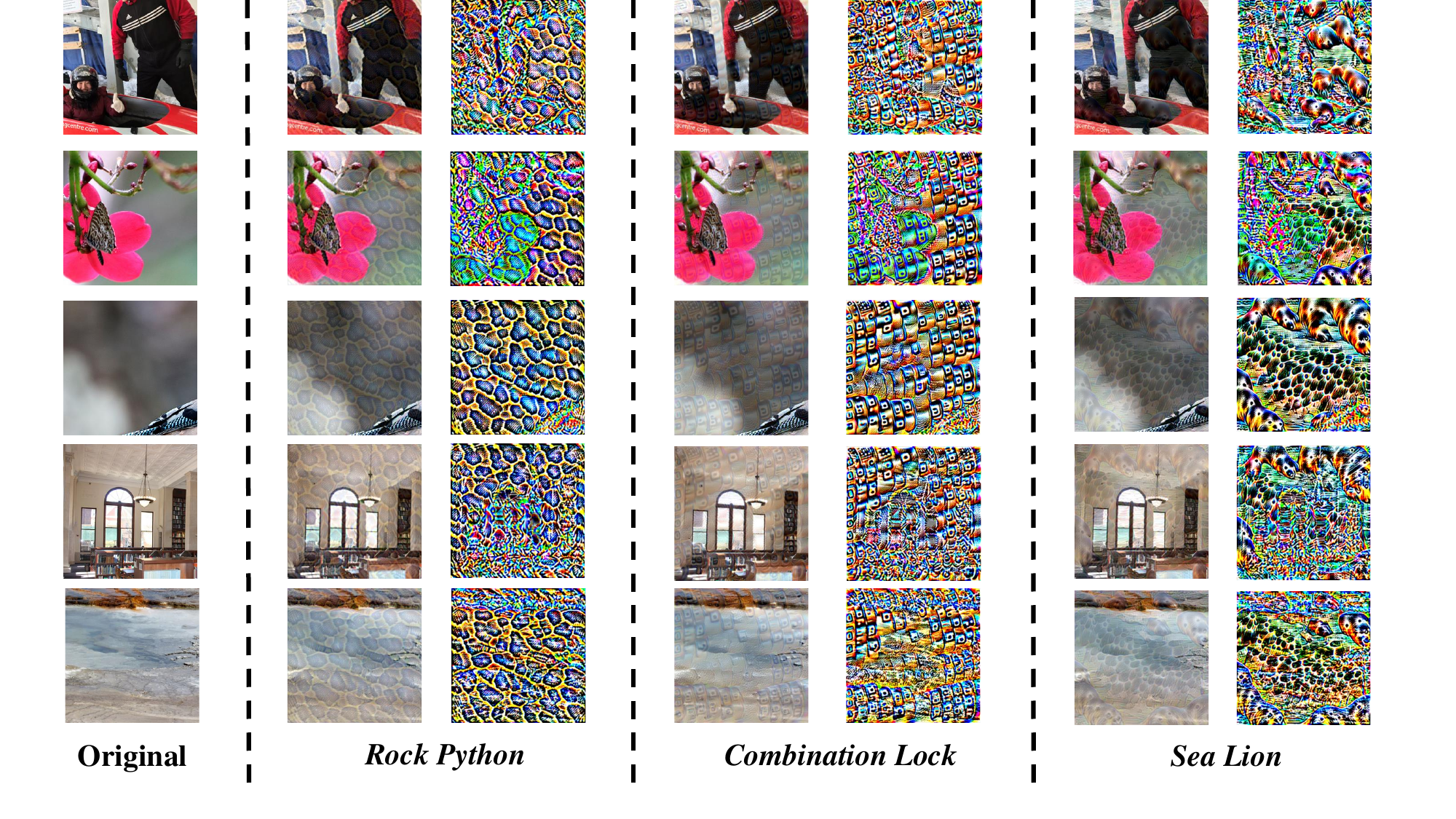}
\caption{
\textbf{Visualization of the adversarial examples with perturbations generated using the proposed TGAF method, employing Inc-v3 as the surrogate model.}
}
\label{fig:app1}
\end{figure*}

\begin{figure*}[!htbp]
\centering
\includegraphics[width=0.9\textwidth]{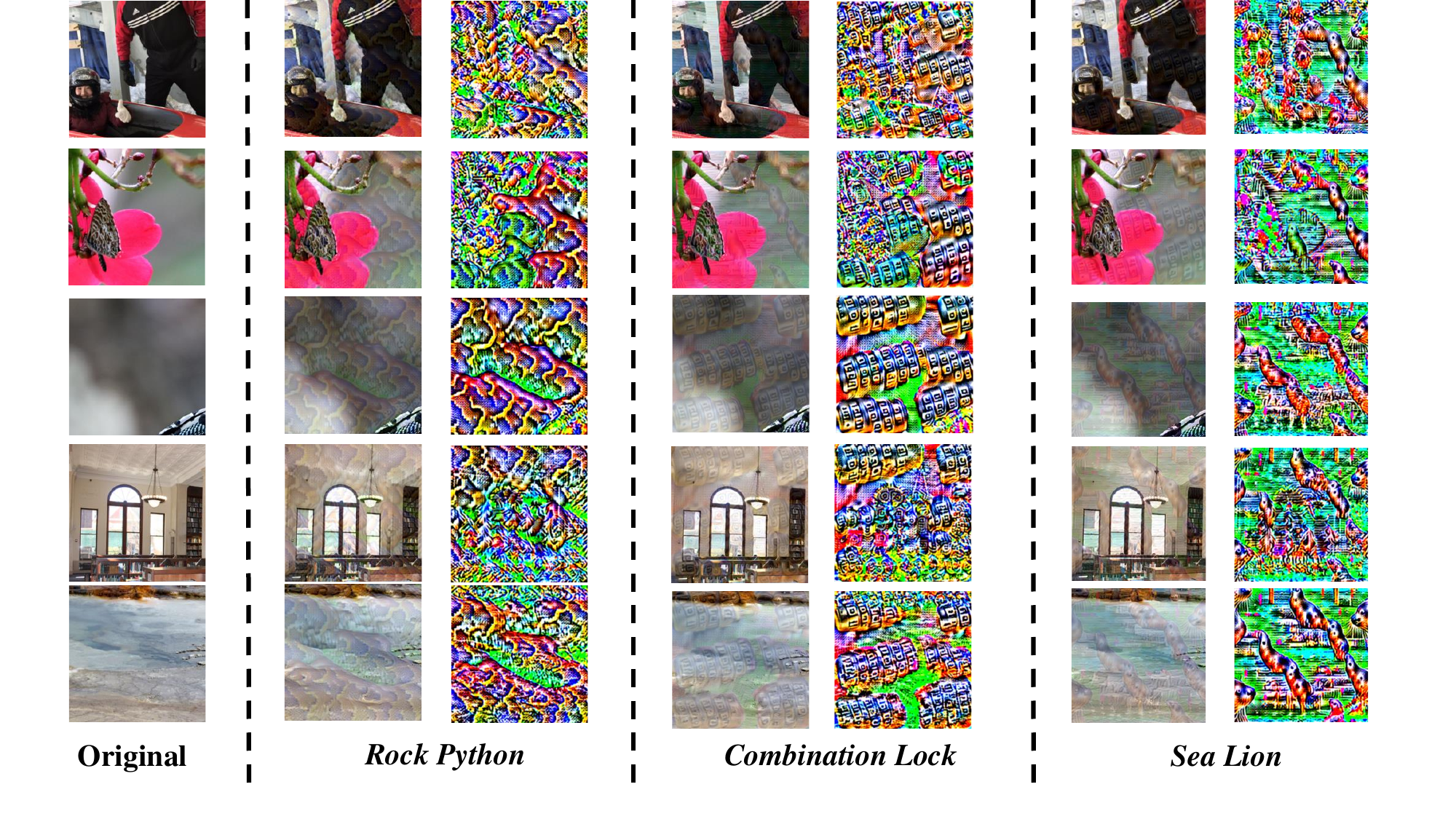}
\caption{
\textbf{Visualization of the adversarial examples with perturbations generated using the proposed TGAF method, employing Res-152 as the surrogate model.}
.}
\label{fig:app3}
\end{figure*}


\section{More Visualization}

In this section, we present additional visualizations of adversarial examples generated by our proposed TGAF framework to further illustrate its effectiveness. Fig.~\ref{fig:app1} displays adversarial examples crafted using Inc-v3 as the surrogate model, while Fig.~\ref{fig:app3} showcases a set of adversarial images produced with Res152 as the surrogate model, alongside their corresponding perturbation maps. The results indicate that TGAF generates a rich and complete target semantic representation.





\section{Implementation Details}

\textbf{2D Tensor Generation and Projection:}
We pre-compute the 2D target tensors using the \texttt{stabilityai/stable-diffusion-2-base} model with a \texttt{DDIMScheduler}. We set the inference steps to \textbf{25}, the CFG scale to \textbf{7.5}, and use a fixed seed (\textbf{seed=0}). This pre-computed $4 \times 64 \times 64$ latent is then passed through our projector, which consists of a 3x3 convolution (stride 1, padding 1) and an \textbf{adaptive average pooling} layer. This projector aligns the $64 \times 64$ spatial dimensions to the $56 \times 56$ feature maps from the image encoder (which expects a $224 \times 224$ input).

\textbf{Masking Mechanism:}
 Our mechanism partitions the image into a 3x3 grid of random-sized, non-overlapping blocks and sets two random blocks to zero. This is applied during training with a fixed probability of 1.0, a strategy validated in Tabs. 5 and 6.




\section{Computational Cost}

As noted in Sec. 4.1, the 2D diffusion latents are precomputed once per target class before training. We also compare the primary computational costs of TGAF against the baseline CGNC in Tab.~\ref{tab:computational_cost}.

\end{document}